\pdfoutput=1

\documentclass[11pt]{article}

\usepackage{acl}
\usepackage[subtle]{savetrees}

\usepackage{times}
\usepackage{latexsym}
\usepackage{amsfonts,amssymb}
\usepackage{amsmath}
\usepackage{mathtools}
    \mathtoolsset{showonlyrefs, showmanualtags, mathic}
\usepackage{mleftright}
\usepackage{subcaption}
\usepackage{xurl}

\DeclarePairedDelimiter\abs{\lvert}{\rvert}

\usepackage[T1]{fontenc}

\usepackage[utf8]{inputenc}

\usepackage{microtype}

\usepackage{inconsolata}

\usepackage{graphicx}

\usepackage{booktabs}

\usepackage{csquotes}
\usepackage{cleveref}
\usepackage{siunitx}
	\DeclareSIUnit{\quantity}{\relax}
	\DeclareSIUnit{\words}{words}
	\DeclareSIUnit{\sentences}{sentences}

\NewDocumentCommand\given{}{\;|\;}
\DeclareMathOperator\softmax{softmax}

\title{Large-scale cloze evaluation reveals that token prediction tasks are neither lexically nor semantically aligned}

\author{
    Cassandra L. Jacobs \\
    Department of Linguistics \\ %
    University at Buffalo \\
    Buffalo, NY, USA \\
    \texttt{cxjacobs@buffalo.edu} \\\And
    Loïc Grobol \\
    MoDyCo\\
    Université Paris Nanterre \\
    Paris, France\\
    \texttt{lgrobol@parisnanterre.fr} \\\And
    Alvin Tsang \\
    Department of Linguistics \\ %
    University at Buffalo \\
    Buffalo, NY, USA \\
    \texttt{alvintsa@buffalo.edu} \\
}

\begin{document}
\maketitle
\begin{abstract}
In this work we compare the generative behavior at the next token prediction level in several language models by comparing them to human productions in the cloze task.
We find that while large models trained for longer are typically better estimators of human productions, but they reliably under-estimate the probabilities of human responses, over-rank rare responses, under-rank top responses, and produce highly distinct semantic spaces.
Altogether, this work demonstrates in a tractable, interpretable domain that LM generations can not be used as replacements of or models of the cloze task.
\end{abstract}

\section{Introduction}

Many language models are trained to a cloze-like objective \cite{taylor1953cloze}, in which they perform next-word prediction (NWP) or masked language modeling (MLM), producing a distribution of probabilities of these tokens given the surrounding context.
Effective language models (LMs) assign the greatest probability to words using the surrounding context.
Humans similarly can perform this cloze task, such as when we try to guess what others are about to say.
By aggregating across many human responses to produce a proportion, one can also obtain a probability distribution over words that fit to a given linguistic context or situation.
Despite major advances in word prediction in context, however, it is generally understood that token prediction tasks cannot fully capture all of the factors that affect human responses, such as real-world knowledge, or syntactic relationships, even when fine tuned \cite{bender2020climbing}.
Cloze productions, while instrumental in understanding language processing, are nevertheless poorly understood as a means of evaluating LM outputs.

Given the continued gap in understanding what makes generated ``language'' natural and human-like, this work aims to better characterize the linguistic-distributional and semantic properties that distinguish LMs from humans in cloze-like tasks.
By better understanding the properties of human productions from a probabilistic perspective, we believe it is possible to bridge the gap between LMs and humans.

Here\footnote{All the code used to produce this analysis is available at \url{https://github.com/calicolab/clamp}.} we compare machine-generated predictions for sentence-final completions against a large-scale human cloze response dataset  \cite{peelle2020completion}.
Such a cloze dataset allows us to specifically target \textit{single-token generations}, which allows us to sidestep the high degree of uncertainty in multiword utterance production, and the many algorithms used to solve these problems \cite{giulianelli-etal-2023-comes}.
That is, not only is multiword utterance production a computationally intractable problem, but different models use different parameters and decoding strategies for the production of longer texts -- effectively dynamical systems -- which makes them difficult to compare and analyze.
Furthermore, selecting high-quality multiword utterances depends on accurately estimating the probabilities of words given the prior context, leading to the explosion of error \cite{Valmari1998}.
Systems that cannot perform single word production cannot be trusted to generate longer, more complex language more accurately.
We therefore present analyses of single-word and single-token response data to provide an estimate of the best case scenario for language modeling.

\section{Human cloze and language modeling}

\begin{table*}
    \centering
    \begin{tabular}{ccccc|c}
        \toprule
        Pythia-70M & Pythia-160M & Pythia-2.8B & RoBERTa & GPT-2 & Human \\
        \midrule
        lot & swarm & swarm & swarm & bee & hive \\
        swarm & bee & bee & bee & swarm & swarm \\
        bee & new & h & hive & hive & bee \\
        new & pest & was & sting & black & nest \\
        threat & h & colony & spider & bad & wasp \\
        \bottomrule
    \end{tabular}
    \caption{Top 5 completions for \enquote{He hated bees and feared encountering a \_\_\_\_}.}
    \label{table:hive}
\end{table*}

The present work examines single-word production in cloze tasks to compare human and LM productions on an identical task.
Note that the term \enquote{cloze} is used in psycholinguistics and natural language processing with close but distinct meanings. 
It always consists of a task setting where an agent (either a human subject or a computer model) is given a context, typically a sentence where a word has been removed, and is asked to complete it with their best guess as to what the original was.
A probability distribution over the possible completions can optionally be deduced from this process. 
However, while these probabilities are meant to estimate some kind of likelihood of occurrence \(\operatorname{P}(w \given c)\) of a word \(w\) in context, they are obtained in significantly different ways.

For the \emph{human} cloze task in psycholinguistics, every human subject in a pool will normally provide one completion \(w\) for every context \(c\) (though see \citealp{roland2012semantic} for an example with multiple responses).
\(\operatorname{P}(w \given c)\) is estimated as the relative frequency of \(w\) among all the completions observed for \(c\). 
For the purposes of simplicity, cloze probabilities are computed under the strong but usually implicit assumption that the choice of a completion by a subject is a non-deterministic process that is identical and identically parameterized across all subjects, so that \(\operatorname{P}(w \given c)\) can  be estimated simply as the prevalence of \(w\) among all responses to \(c\) \citep{smith2011cloze}.

For \emph{language models}, and in particular for recent \emph{neural} ones (NLM), the cloze task is cast as classification task, where the inputs are the contexts, the classes are the words in a vocabulary, and the classification is done by assigning a score to each possible word using a neural network and choosing the one with the maximal score. 
A probability distribution over words can be obtained by applying the \(\softmax\) function to the output layer of this neural network, which is usually trained by using a variant of the stochastic gradient descent algorithm to minimize the negative log-likelihood of the removed word for every context in a large training corpus. 
While there is no guarantee of it\footnote{Neither theoretically nor empirically, especially without explicitly optimizing for it --- see e.g.\ \citet{kong-etal-2020-calibrated} or \citet{ulmer2024calibratinglargelanguagemodels}}, it is often\footnote{This is for instance the assumption made by \citet{Radford2019LanguageMA} and \citet{devlin-etal-2019-bert}.} assumed that this probability distribution is learned to fit a latent likelihood of occurrence of a word in context.

Cloze data are significantly more sparse than language model predictions. 
Even the largest cloze datasets collect \num{40} to \num{100} human responses per context only and the number of \emph{unique} responses is often much lower \citep{lowder2018lexical, luke2018provo, peelle2020completion}. %
NLMs, on the other hand, typically generate predictions for over tens of thousands of tokens at once.
It is precisely this estimation of the \enquote{long tail} of uncommon or improbable words that has been argued to account for the majority of their strength over human cloze data  \cite{szewczyk2022context,shain2024large,oh2023does}.

Under these assumptions, the probability distributions obtained from a NLM (provided that a suitable one exists) should be highly correlated to those obtained from human cloze experiments and easier to obtain, leading to their use as a proxy in psycholinguistics works \cite{luke2018provo,staub2015influence,de2024cloze}. 
Conversely, human cloze datasets can serve as benchmarks for NLMs: if it does indeed learn human-like probability distributions, a good NLM should provide distributions that are highly correlated to those obtained from human cloze data.

\section{Data}

We experiment on \citeposs{peelle2020completion} completion norms dataset. It consists of \num{3085} English sentences for which human participants were asked to provide a final word, with each sentence receiving at least \num{100} manually validated responses in order produce reliable cloze probability estimates.
The stimuli varied between eight and ten words in length, and varied in the degree of final-word predictability.
The sentences are importantly long enough that LMs should be able to reliably guess appropriate words for the context.

We compare the cloze probabilities reported by \citet{peelle2020completion} to probabilities extracted from several neural language models: GPT-2 \citep{Radford2019LanguageMA}, RoBERTa \citep{Liu2019RoBERTaAR} in its \enquote{base} configuration, and the Pythia suite of language models \citep{biderman2023pythia}. 
All of these models use a subword vocabulary size of approximately \num{50000} and thus have the same capacity for long tail effects, though they differ in their training datasets and subword vocabularies.

In contrast to GPT-2 and RoBERTa, the Pythia models provide a way of exploring the influence of several hyperparameters on their behavior, namely model size, training time, and data deduplication. 
Pythia models are trained to perform NWP on the Pile \citep{gao2020pile} using a training procedure similar to GPT-2 and have been used extensively to evaluate LM fit to human reading data \cite{oh-etal-2024-frequency,oh-schuler-2023-surprisal}.
These models are also interesting because their embedding (first) and projection (last) matrices are untied, in contrast to other models such as GPT-2, which could in principle produce different kinds of predictions.
We note that due to their more recent release and because they were not designed for downstream task performance such as natural language inference, Pythia models could provide a poor baseline that will not generalize well to non-language modeling tasks (although the evaluations provided by \citet{biderman2023pythia} show that it is in fact not the case).
In Experiments \href{sec:exp-proba}{1} and \href{sec:exp-rank}{2}, we include only Pythia-160M because it contains a similar number of parameters as GPT-2 and RoBERTa and because the highest-performing, largest models are impractical for quick inspection of their behavior.
However, we assess the role of model capacity in \nameref{sec:exp-pythia}, which leverages a wider range of Pythia models to study the influence of model size and training budget.

We also assess GPT-2 (a decoder-only NWP model) and RoBERTa (an encoder-only MLM), which produce representations that have improved performance on many downstream NLP tasks and are perhaps the best-known examples of their respective flavor of Transformer-based LMs. 
Examining these earliest Transformer LMs ensures that the effects we observe are not tied to a possible low quality of the models in general: it has been empirically shown that they are good at what they have been designed for.

In all our experiments on model generations, we only consider predictions of a single word-initial subword token, which can either be a \enquote{word} or a fragment. 
While this can be somewhat limiting in theory because sometimes a human response might consist of several subword tokens\footnote{In practice, we also found that most human responses were common words that are only a single token in the vocabularies of the models we study, so the distortion to the results we observe is generally negligible.}, this decision makes it possible to directly estimate ranks and probabilities without having to resort to a sampling procedure \citep{holtzman2021surface, giulianelli-etal-2023-comes}, which would otherwise make our results less interpretable. 
Directly measuring probabilities across the full sequence of generations is already a critical component of these sampling algorithms, but the relative computational complexity of generation of single tokens is much lower, enabling further experiments.

\section{Experiment 1}\label{sec:exp-proba}

\begin{figure}[th]
    \centering
    \includegraphics[width=.9\linewidth]{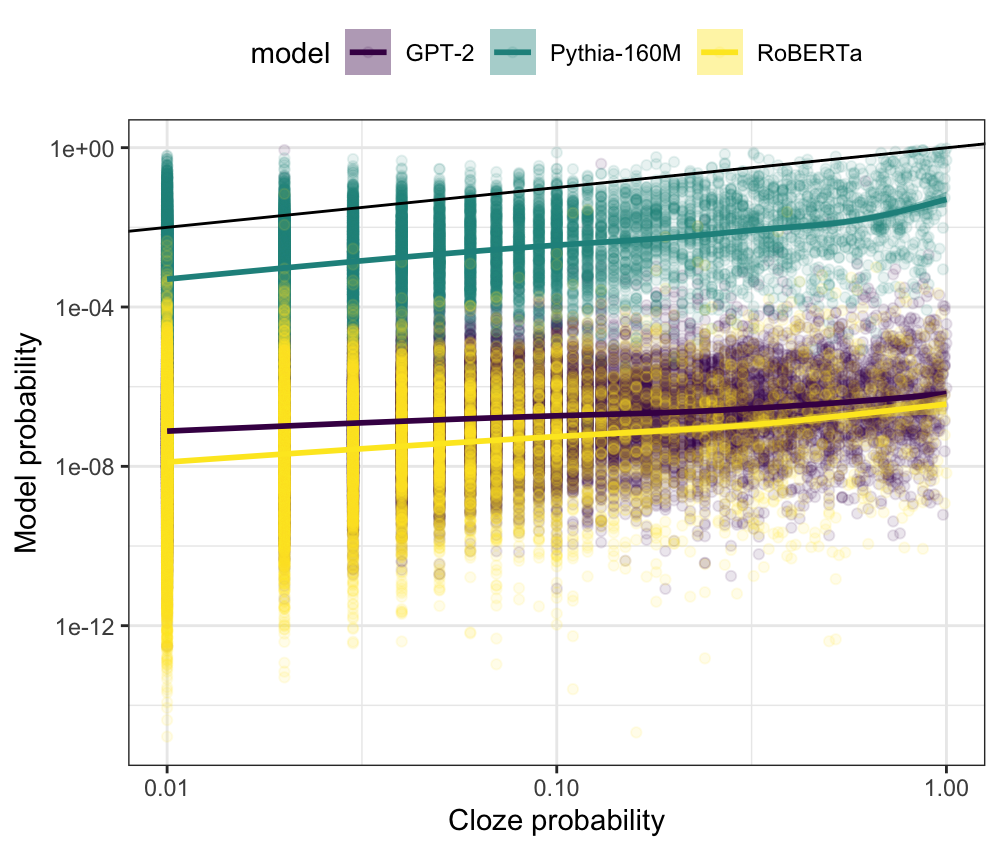}
    \caption{Correlation between human cloze probabilities and language model probabilities showing clear non-linearity in correlation and massive under-estimation of next-word probabilities.}
    \label{fig:cloze-probability-corr-plot}
\end{figure}

This experiment compares GPT-2 \citep{Radford2019LanguageMA}, RoBERTa \citep{Liu2019RoBERTaAR}, and Pythia \citep{biderman2023pythia} to understand the probabilities of next-word generations by comparing human- and LM-derived estimates.
Qualitatively, the generations of RoBERTa, GPT-2, and Pythia-160M vary widely in the content of their predictions. We present one example in Table \ref{table:hive}.

In order to be used as models of human productions, the top LM predictions and/or their estimated probabilities should at the very least be consistent with observed human responses to the same cloze challenge.
However, Figure \ref{fig:cloze-probability-corr-plot} illustrates that LMs most often assign too little probability to human generations \citep{holtzman2021surface}, with most models under-estimating cloze probabilities by several orders of magnitude.

These results also reveal that the item-level correspondence between LM and human cloze probabilities is poor, with considerable variance around the trendline and evidence of non-normality \citep{franke2024bayesian}.
Taken together, these findings suggest that LMs are generally \enquote{right for the wrong reasons} with respect to questions of predictive power \citep{giulianelli2024generalized,shain2024large}. %

We note that the probabilities and correlations extracted from GPT-2 and RoBERTa are much lower than previously reported in other works \cite{frisch-giulianelli-2024-llm,oh2024leading}.
One possibility is that the reliance on web text is a poor estimator of human cloze probabilities \citep{smith2011cloze}.
For instance, the subword vocabularies learned by RoBERTa and GPT-2 may be skewed in favor of decomposing responses into multiple subwords in human cloze data.
By contrast, the Pile \citep{gao2020pile}, which was used to train Pythia, contains much more diverse texts that are more representative of everyday language use, and may produce estimates of next word probabilities that are closer to psycholinguistic stimuli.
Another possibility is that the differences in stimuli used by different studies (e.g., \citealp{de2024cloze}) produces different distributions of estimates. 
Finally, ongoing work suggests that subword aggregation plays a significant role \cite{nair-resnik-2023-words,oh2024leading} in the estimation of next-word probabilities.

\section{Experiment 2}\label{sec:exp-rank}

\begin{figure}[th]
    \centering
    \includegraphics[width=0.9\linewidth]{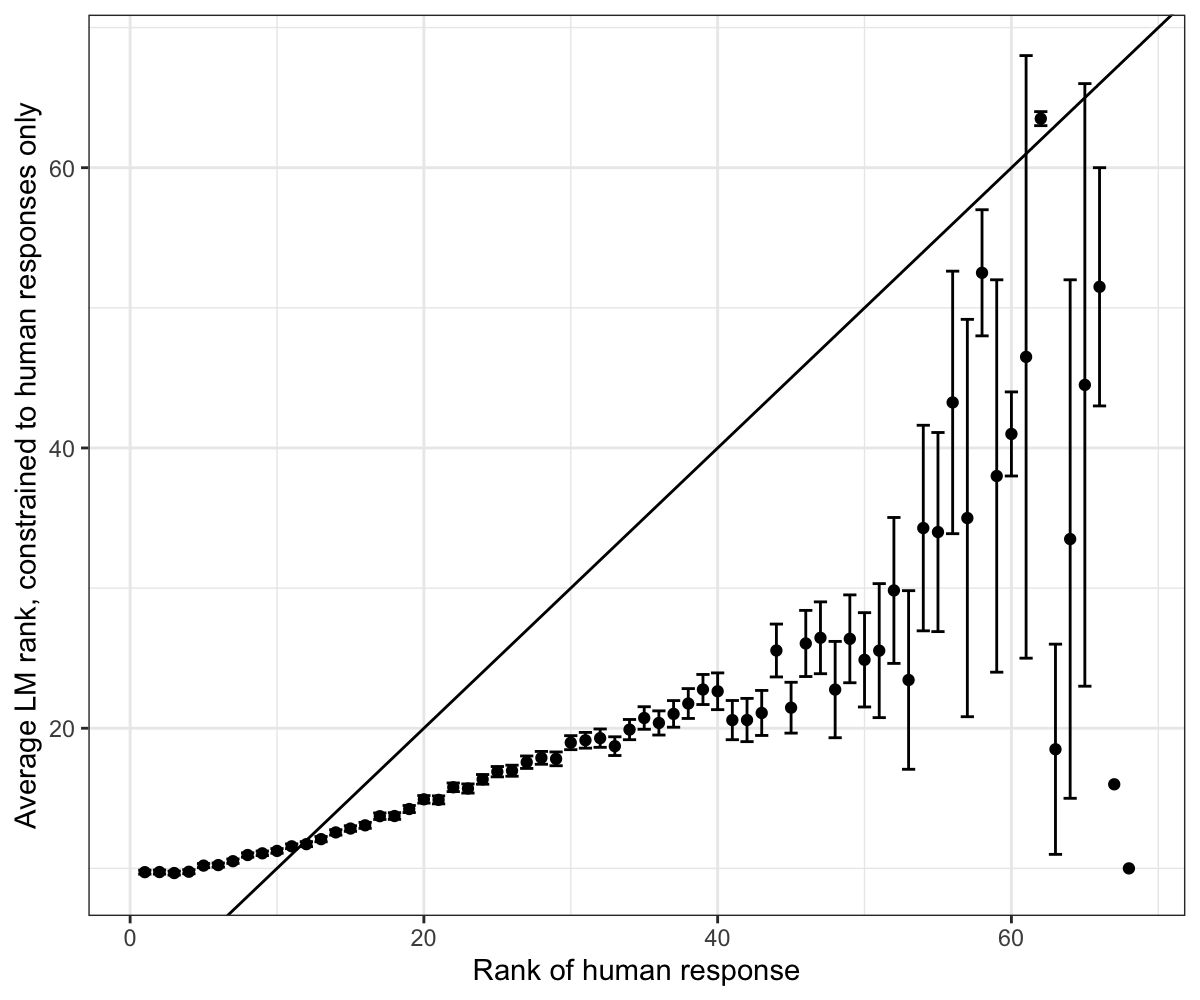}
    \caption{Rank correlation between Pythia-160M and human responses. Language models over-rank rare human responses (above solid line) and under-rank probable responses (below solid line).}
    \label{fig:rank-corr-plot}
\end{figure}

\nameref{sec:exp-proba} identified significant problems in the use of LM-based probabilities, but the selection of next words is largely one of ranking different candidates.
One could reason that relying only on probabilities is unfair to language models, with generally reasonable rankings but poor estimation of the probability of next words due to data or training objective differents.
For simplicity, we examined whether the Pythia-160M model (most similar in size to the state-of-the-art GPT-2; \citealt{biderman2023pythia,shain2024large}) was also miscalibrated in its rankings of human responses by comparing the rank of the probability associated with the first subword of a human response versus the empirical rank in the cloze dataset.
This allows us to constrain the correlation to only valid human responses and determine if the language models are able to accurately order them, as in a multiple-choice test.

Figure \ref{fig:rank-corr-plot} shows that one need not be optimistic about ranks here either.
The rank correspondence between language model probability and human cloze probability suggests that Pythia-160M consistently under-estimates the ranks of probable responses (left), and over-estimates the ranks of rarer responses (right).

\begin{figure}[th]
    \centering
    \includegraphics[width=0.9\linewidth]{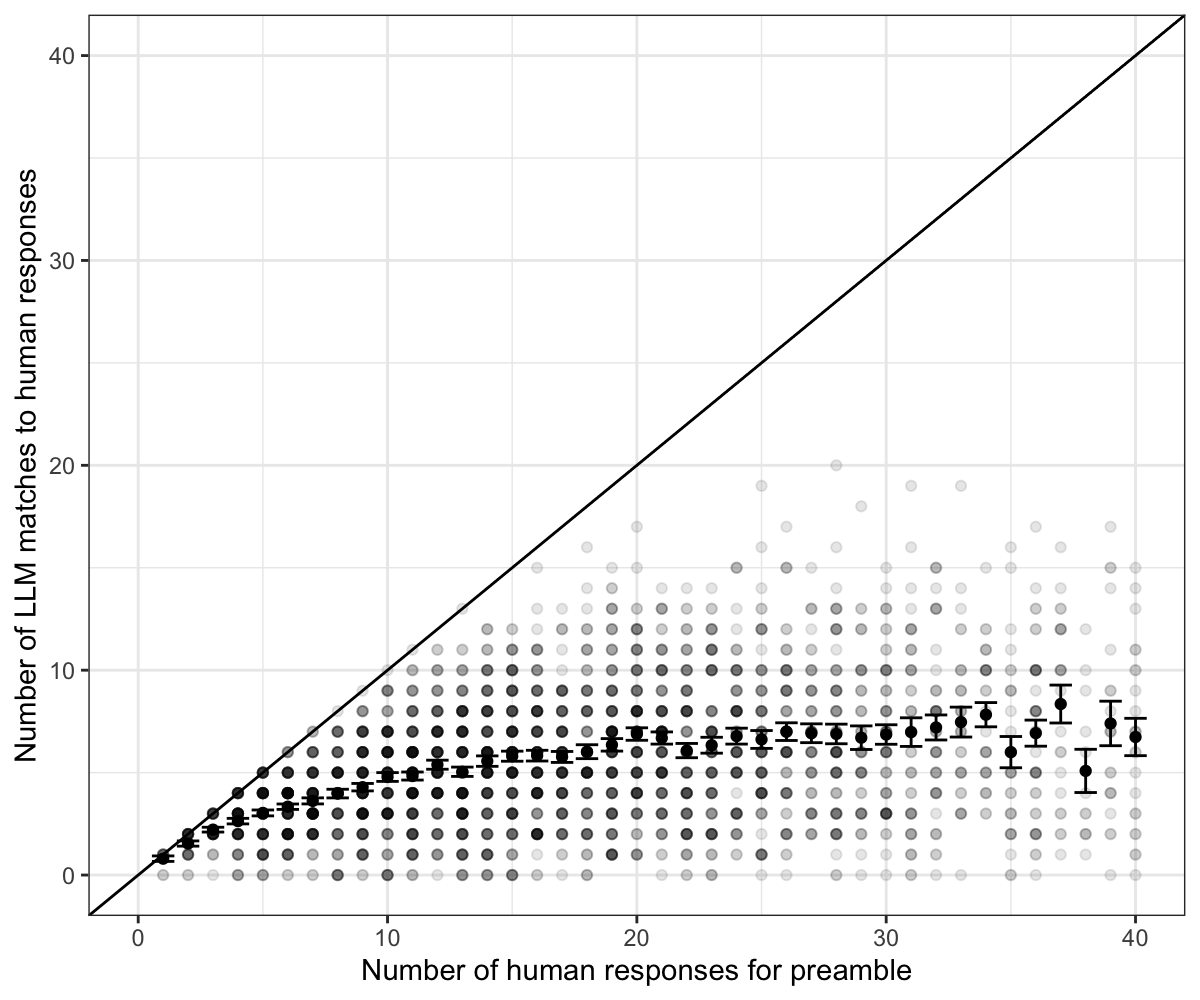}
    \caption{Pythia-160M next word prediction recovery compared to human productions.}
    \label{fig:count-recall}
\end{figure}

\begin{figure*}
    \centering
    \includegraphics[width=.8\linewidth]{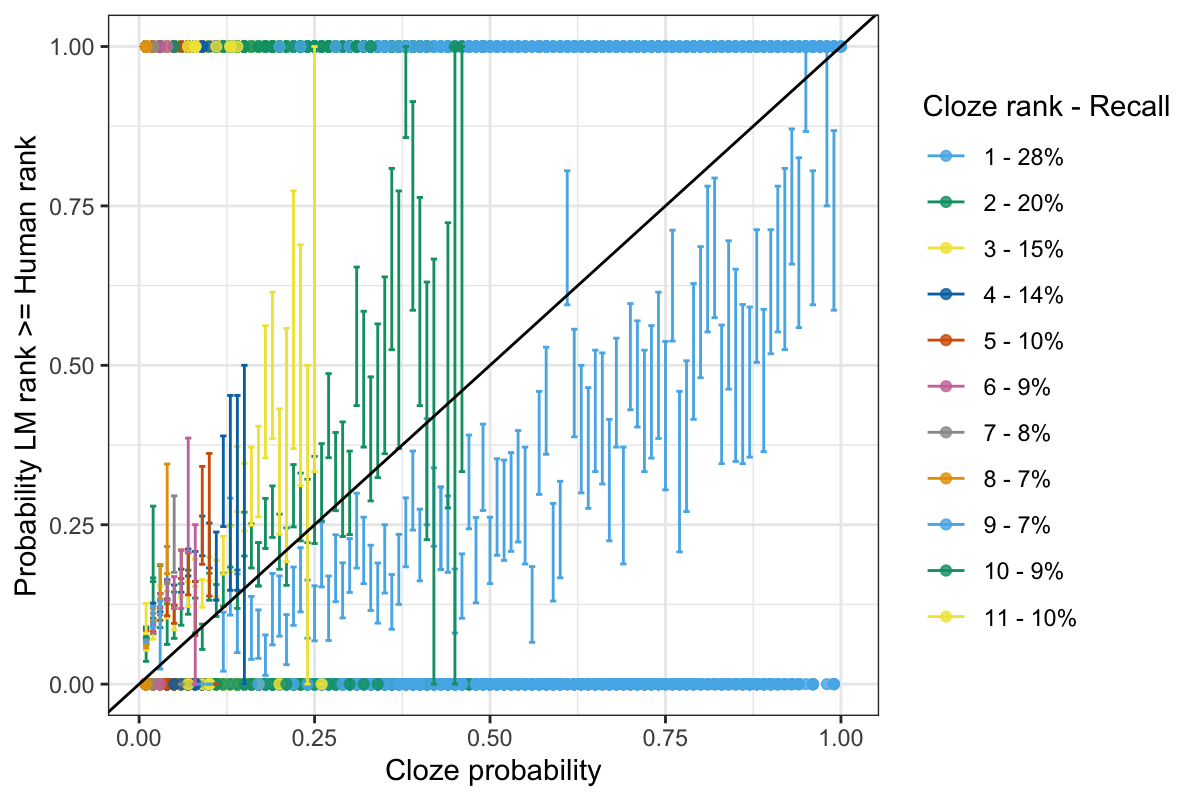}
    \caption{Pythia-160M recovery by cloze response rank and proportion of recoveries. Note that no responses with rank 11 are recovered at an equal or higher rank.}
    \label{fig:proportion-recall}
\end{figure*}

An additional perspective on this ranking problem requires inspection of precision and recovery of cloze responses.
We present in Figure \ref{fig:count-recall} the recovery of human responses that appear in the top \num{40} predictions of Pythia-160M.
Generally speaking, when the LM is constrained to produce approximately the same number of unique responses as humans could provide in the cloze task, it is unable to recover more than \num{10} of the human responses in the top \num{40} responses.
Indeed, per this Figure, it is clear that the proportion of responses would fall and a larger threshold would be necessary to recover rare human responses.

However, it is possible that top human responses are simply poorly ordered within the models' top responses.
We examine whether a more lenient criterion would improve fit to human cloze production
by allowing ranks to include any higher ranks.
Figure \ref{fig:proportion-recall} shows the performance of this model on recovering words at the correct ranking or higher. 
It is clear that only very probable responses (i.e.\ sentences that produce only one or two unique completions) are recovered by this model, and that human and LM predictions diverge rapidly in more uncertain contexts.
Even then, top responses are retrieved in first place by the models only \qty{28}{\percent} of the time, while rarer responses in the top 10 human responses are selected at that rank or higher only \qty{9}{\percent} of the time.
Strikingly, this is an over-estimate, such that second-best and lower answers answers in the human data appear to be commonly ranked as the top human response.
Indeed, the ranks of top human responses seem to be cannibalized by the higher-ranking behavior of lower-probability human responses.

\section{Experiment 3}\label{sec:exp-pythia}

\begin{figure*}
    \centering
    \begin{subfigure}[t]{0.49\textwidth}
        \centering
        \includegraphics[width=.95\linewidth]{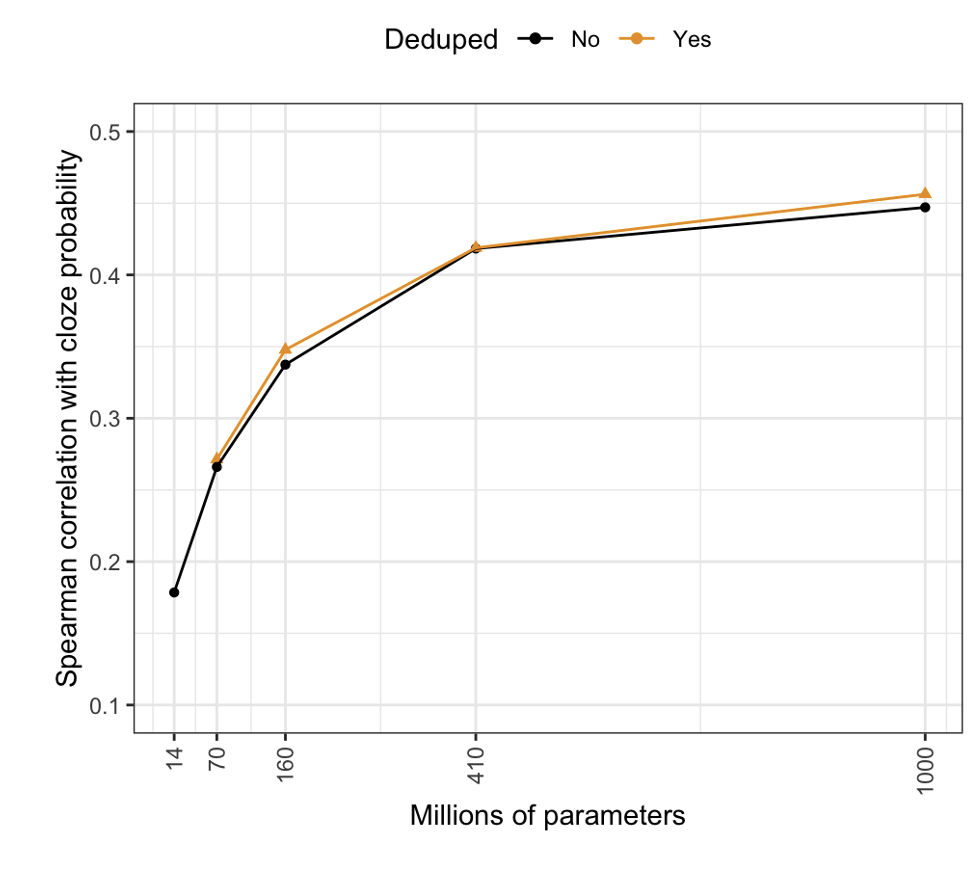}
        \caption{Effect of model size on rank correlations with human responses.}
        \label{fig:pythia-nparams-spearman}
    \end{subfigure}
    ~
    \begin{subfigure}[t]{0.49\textwidth}
        \centering
        \includegraphics[width=.975\linewidth]{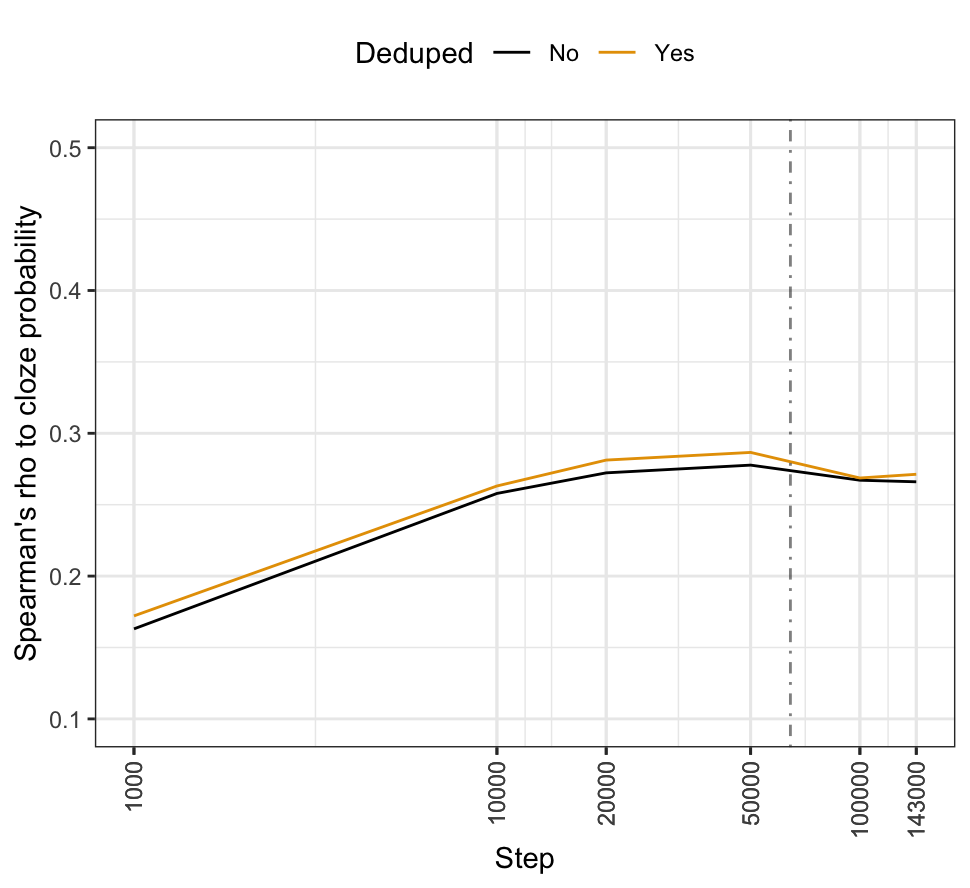}
        \caption{Effect of training budget on rank correlations with human responses.}
        \label{fig:pythia-nsteps-spearman}
    \end{subfigure}
    \caption{Spearman rank correlation between LM (Pythia 160m-deduped) and human responses as functions of model size and training budget. Both show positive correlation plateauing after a certain point, staying at \(\rho<0.5\) out of a maximum of \(1\).}\label{fig:pythia-correlations}
\end{figure*}

This experiment assesses the contribution of learning in LMs to fit to human data. 
To assess the influence of model size and training progress on the correlation between NLM and cloze rank distributions, we experiment with the Pythia suite \citep{biderman2023pythia}, a series of Transformer-based NLM spanning a wide range of model sizes for a single architecture, trained on the same data and with publicly available intermediate checkpoints. 
For comparisons based on model size, we use all the models up to and including Pythia-2.8B\footnote{The lower bound for the \enquote{phase change} in model behavior observed during training by \citet{biderman2023pythia}.}, in standard and \enquote{deduped} versions both.
For comparisons based on training progress, we use checkpoints of Pythia-160M\footnote{According to Figure \ref{fig:pythia-nparams-spearman}, the 2.8B version would have had a slightly better correlation overall, but to reduce the cost of our experiments, and because given the results reported by \citet{biderman2023pythia} we expect the trends relatively to training budget to be similar enough, we only experiment with the 160M checkpoints for Figure \ref{fig:pythia-nsteps-spearman}.} at five training steps from \num{1000} to \num{143000}.

We measure the similarity between human and LM rank distributions with \citeposs{spearman1904ProofMeasurementAssociation} rank correlation coefficient \(\rho\).
For LMs, we only consider the relative ranks they give to human responses and ignore all preceding and intervening responses. In other words, these correlations only answer the question \enquote{Did the LM rank human responses in the same order humans did?}. 
As in \nameref{sec:exp-proba}, this allows us to obviate any discrepancy due to the long tail of LM vs.\ human responses, but it may present an optimistic estimate of the fit of the LM to human production data.

LM performance generally improves logarithmically with model size, such that greater model sizes have higher correlations with human responses, but shows clear diminishing returns in terms of computational budget.
Additionally, deduplicated text typically produces higher-quality estimates of cloze productions, which suggests that highly repetitive data sources such as web text will consistently under-estimate human language productions \cite{oh2023does,smith2011cloze}.
Surprisingly, the performance of the five checkpoints we examined reveals diminishing returns over the course of multiple epochs as well, suggesting that more training data does not necessarily lead to better estimation of human linguistic creativity.

Figure \ref{fig:pythia-nparams-spearman} shows a trend of steady increasing of correlation between ranks as model size increases. However, even in this lenient setting, the largest models stay at \(\rho<0.5\), an at best moderate correlation, and the trend suggests rapidly diminishing returns over \num{410} million parameters. 
As for training steps, Figure \ref{fig:pythia-nsteps-spearman} shows that the correlation between human and LM ranks --- while overall mediocre --- also increase during training, but plateaus around \num{50000} steps.

\citet{biderman2023pythia} report that model performance is positively correlated with both model size and training time. Our results show the same trend for correlation with human responses. The drop in performance at longer training budgets replicates some prior work suggesting that the number of tokens can influence predictive power \cite{oh2023does}.
However, the generally low correlations at all sizes and all training times suggest that even if these trends persisted, it would be impractical to rely solely on them to obtain good models of human productions.

\begin{figure}
    \centering
    \includegraphics[width=1\linewidth]{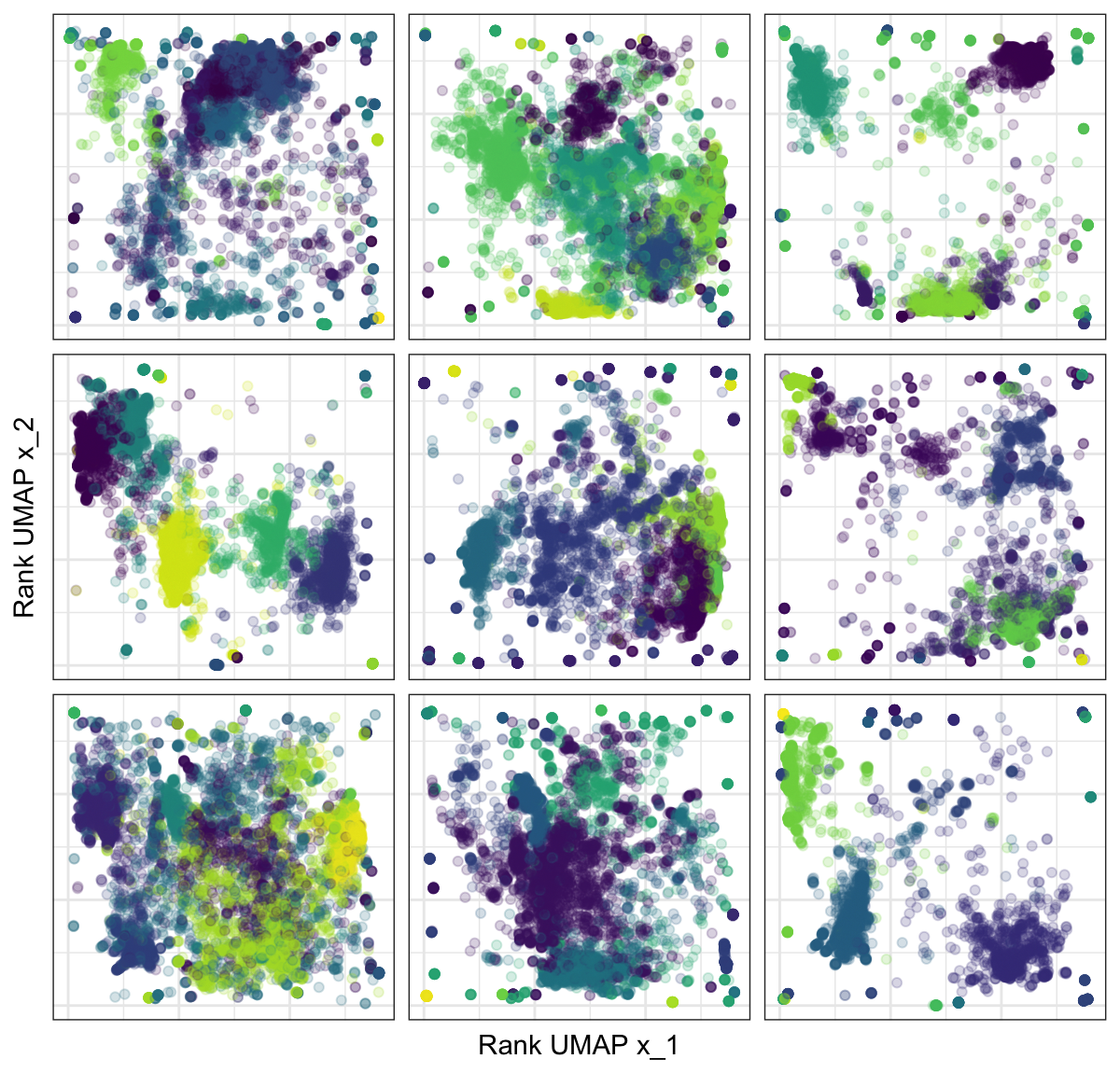}
    \caption{UMAP demonstrates high degrees of cluster separability in human-generated productions.}
    \label{fig:pythia-umap}
\end{figure}

\section{Experiment 4}\label{sec:exp-clusters}

The results of Experiments \href{sec:exp-proba}{1} and \href{sec:exp-rank}{2} showed modest correlations between human- and machine-generated responses in a cloze task for approximating the probabilities and ranks of human responses, replicating previous work showing deficiencies in next-word prediction \cite{holtzman2019curious,vaidya2023humans,botch2024humans,klein2024effect}.
Moreover, the improvements in fit to cloze probabilities from greater model size and longer model training seem modest, suggesting that better language model memorization comes with poor creativity gains.
That is, language models show clear effects of over-fitting to the training data \cite{shain2024large,oh-etal-2024-frequency}.
While this and previous results underscore how clearly language models and humans prefer to produce responses in a different order from each other, it is possible that human and language model responses are more closely linked on a semantic level due to the distributional training regime \cite{baroni2014don}.
Furthermore, individual responses may effectively express the same meaning while being written slightly differently \cite{holtzman2021surface}.
Thus, this experiment examines the distributional semantic properties of human and language model responses and assesses the degree to which they inhabit the same geometric space.

To better understand the semantic structure of generations by humans and language models, we perform clustering experiments on the middle layer (layer 7) from the Pythia-160m-deduped model.
We selected this layer due to the apparent greater compositional capacity of the middle layers of language models \citep{tenney-etal-2019-bert, jawahar2019does, antonello2024predictive}, which are less tuned toward next-word prediction and better reflect a word's role in a sentence.
To cluster the responses, we use the scikit-learn \citep{scikit-learn} implementation of the Bayesian Gaussian mixture model (BGMM), which uses a Bayesian Dirichlet prior \cite{blei2001latent} to approximate the distribution of \(n\)-dimensional embeddings of human and/or model responses as samples from a mixture of \(n\)-dimensional multivariate normal distributions.

We are inspired by recent work that has applied BGMMs to learn coherent cluster representations with relatively stable parameters for cloze data \cite{jacobsuncovering}, which we apply here.
A visual inspection of the human responses, which we plot in Figure \ref{fig:pythia-umap} using UMAP over pairwise Euclidean distances between the embeddings \cite{mcinnes2018umap}, shows a high degree of local clustering.
Data points, which are colored according to the Bayesian Gaussian mixture model's cluster assignments after training, show considerable separation between the different clusters, demonstrating the utility of this clustering algorithm for interpretation.

If human and LM generations have different semantic network topologies, then we expect the degree of human-versus-LM content in a dataset to influence the clusterings that are obtained, both qualitatively and quantitatively.
For this experiment we construct four BGMM models trained on embeddings belonging to different categories of generation.
We ultimately wish to assign cluster labels to three response-based categories, which are human-only responses (\textsc{human}; \num{39995} tokens), LM-only responses (\textsc{lm}; \num{106253} tokens), and shared responses (\textsc{both}; \num{17147} tokens). 
To minimize the effect of dataset size on clustering performance, we constructed clustering models that sampled the input to the mixture model to have the same number of data points as the human-only dataset.

The results of the clustering analysis provide evidence that responses that are shared by both the Pythia model and human participants are unique from other types of responses.
In terms of clustering model likelihoods assigned to out-of-sample data, the human-only and LM-only clustering models have poor clustering performance on responses produced by both generative processes (Table \ref{tab:likelihoods}, but a model that is designed to cluster those responses shows a high degree of coherence relative to all other models, and is in fact the only clustering model with a positive mean likelihood.
Additionally, it is clear that human responses are also poorly explained by language model representations, with substantially lower likelihoods than LM-based likelihoods.
One way to interpret these data is that LM-based clusters are of sufficiently high quality to assign LM-generated next token predictions, but the resulting representations for human productions are of lower quality due to being out-of-distribution.

We next compare the similarity of the cluster assignments across different clustering datasets.
Specifically, we compute a measure known as Clustering Mutual Information (CMI), a special case of the general Clustering Agreement Index (CAI; \citealp{Rabbany_Zaïane_2017}) that estimates the degree of overlap between multiple clustering models (i.e., across different LM implementations) for the same data points (productions).
We review the method in depth in the Appendix. 
Table \ref{tab:CMI} shows the clustering agreement across each of the Gaussian mixture models that we trained.
The human-derived clustering model provided the highest degree of agreement with responses produced by both the human and Pythia-160M-deduped model, though the LM-only clustering model proved similarly effective.
On the other hand, the LM-only model and the model trained on shared human and LM responses had relatively poor performance on human data.
Note that all models have low degrees of agreement in general relative to the ideal scenario.

\begin{table}
    \centering
    \begin{tabular}{lrrr}%
         & \textsc{human} & \textsc{lm} & \textsc{both}\\ %
         \hline
      \textsc{human}   & \textbf{-1.79} & -3.95 & -6.60 \\
      \textsc{lm}   & -1.95 & \textbf{-0.93} & -2.18 \\
      \textsc{both}   & -8.55 & -7.21 & \textbf{0.10}\\
    \hline
    \end{tabular}
    \caption{Mean data likelihood scores per observation for different data sources (rows) and mixture models (columns).}
    \label{tab:likelihoods}
\end{table}

\begin{table}
    \centering
    \begin{tabular}{lrrr}%
         & \textsc{human} & \textsc{lm} & \textsc{both} \\
         \hline
        \textsc{human} & - & \textit{.478} & \textit{.479} \\
       \textsc{lm}  & .505 & - & \textbf{.519} \\
       \textsc{both}  & \textbf{.600}  & .596 & - \\
       \hline
    \end{tabular}
    \caption{Clustering agreement indices for different data sources (rows) and mixture models (columns).}
    \label{tab:CMI}
\end{table}

\section{Related work}
Our approach is inspired in part by \citet{smith2011cloze}, who explored the differences between n-gram language models and human productions in a next-word cloze task. Their study found that relative to 5-gram probabilities drawn from web text, human responses in the cloze task were more concrete, familiar, diverse, and semantically related to the preceding context. Their work suggests that cloze productions deviate substantially from web text in ways that should be evident in our experiments.

Relative to human predictions for sentence-final completion data, NWP models such as GPT-\{2, 3, J\} appear to produce better estimates of the predictability of a word, with evidence mostly coming from correlational studies of large-scale reading time datasets \cite{shain2024large}.
However, recent studies have repeatedly shown that linguistic predictability quantified by language models lags behind other sources of linguistic knowledge.
Moreover, while the predictive power of modern LMs is a significant improvement beyond n-gram language models \cite{goodkind2018predictive}, modern language models are commonly miscalibrated to most tasks, such as natural language inference.

The reasons for the lack of predictive power of human cloze data are still largely unknown, but are generally attributed to the expense of probing enough humans in enough linguistic contexts to accurately measure next-word probability.
However, it seems clear that at present, LLMs cannot function as a drop-in replacement for cloze estimates.
These issues present a challenge for computational psycholinguistics about the claim that LM predictions are fundamentally better for capturing human predictions \cite{shain2024large}.
off-the-shelf language models must still be calibrated toward human language prediction tasks.

The model capacity experiments we present build on some previous work \cite{oh-schuler-2023-transformer}, which reported a peak in fit to reading time data at approximately step 1000 in training the same class of Pythia models, which is earlier than we observe in our analyses.
The effect that is seen in this literature may arise due to differences in sensitivity to word frequency depending on the psycholinguistic task \cite{oh-etal-2024-frequency,klein2024effect}.

Adjusting the temperature of the \(\softmax\) function when computing probabilities from LLMs' outputs has been found to improve predictive power for reading times \cite{liu2023improving}.
We note that temper does not affect the ranks of responses, unless temperature settings are so high as to collapse all low-probability responses to the same value, and so temperature scaling would not affect massive rank disparities relative to human cloze probabilities.

Previous clustering approaches other than Gaussian mixture models have been applied to improve intepretability of LM outputs \cite{thompson2020topic,grootendorst2022bertopic} and have demonstrated that some LMs cluster contextual embeddings more effectively than others.
However, this work has not examined human or LM productions together.

\section{Discussion}

Given the differences for participant responses against human ranks in both LM ranks and LM probabilities, it is clear that LMs' ability to directly model human production behavior is quite limited, even at the single-token level.
These observations are consistent across models with different training data (Pythia vs GPT-2) and training objective (RoBERTa's MLM vs.\ Pythia and GPT-2 NWP). 
The fact that neither ranks nor top-k prediction sets match properly suggests further that these divergences are not due to probability distortions due to long tail/truncation effects.

In addition to poor probabilistic and ranking behavior, the semantic spaces generated by the model that we examined are not semantically aligned with human productions.
Human beings' responses are poorly clustered by mixture models trained on non-human generations, and the types of responses that humans and LMs produce similarly poorly capture the semantic diversity of human responses that LMs do not predict.

The similarity between human cloze results and LMs are dissimilar enough that LMs cannot be used as proxies for the human production process.
We believe that these results present a cautionary tale, in which poor fit of cloze probabilities to reading time data relative to language model probabilities arises not from sparsity but the inability to retrieve human-like predictions, as evidenced by weak correlations and substantial under-estimation of human responses.
This suggests that the use of LM probability estimates may not rely on the same types of predictive processes at all, counter to much recent work \cite{shain2024large,blank2023large}, and that researchers should be very careful in general when using LMs as proxies for any predictive modeling.

\newpage

\section{Limitations}

In this work, we evaluate only a single dataset to understand the correspondence between cloze probabilities and language model estimates, which may limit the ability to draw conclusions about the nature of the relationship for all cloze tasks and datasets.
Some of the cloze datasets that we did not examine include word-by-word cloze completions that allow researchers to assess the probabilities of words at each stage in sentence processing.
Additionally, different subword tokenization strategies can have an impact on estimates, and our approach to select only word-initial subwords may inflate estimates of next-word probabilities \cite{oh2024leading}.
Finally, it is common in the literature to report $\Delta$ log likelihood instead of correlation coefficients; we do not report this here because we do not do model comparison, as we are concerned entirely with unbiased generations.
However, future work should compare and contrast these approaches directly.

\section{Ethical Considerations}
The \citet{peelle2020completion} data were gathered using crowdsourcing on Mechanical Turk. The original researchers obtained IRB approval for their research study. Some of the human responses, and many of the LM predictions, contain profanity, sexually explicit content, or other offensive content. 
Furthermore, LMs in general are likely to produce sexist, ableist, and racist responses even when guardrails are implemented. Researchers seeking to evaluate LM outputs using human judgments should be aware of the potential for harmful material in these outputs.

\bibliography{anthology,custom}

\appendix

\section*{Pearson correlation}

\begin{figure}[th]
    \centering
    \includegraphics[width=1\linewidth]{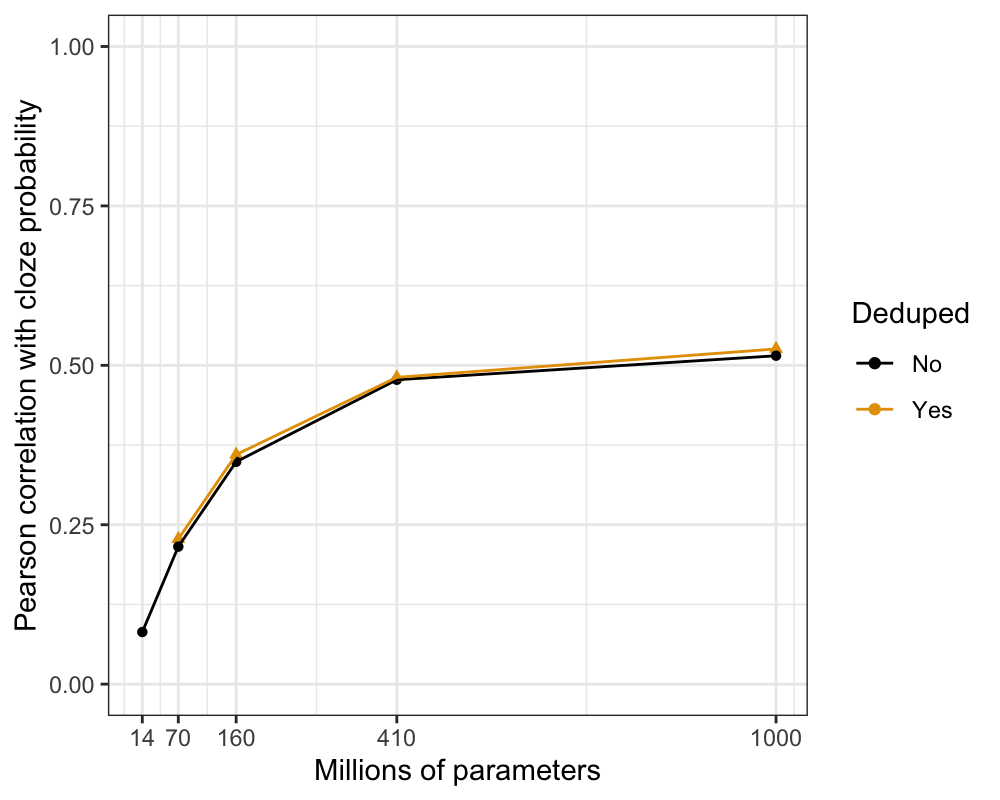}
    \caption{Pearson correlation between model capacity and rank correlations with human responses. Pearson correlation is significantly higher than Spearman correlation for larger model sizes and signifantly lower for smaller models.}
    \label{fig:pythia-nparams-pearson}
\end{figure}

\section*{Clustering Agreement Index (CAI)}

A CAI is utilized to generate a numerical value that portrays the similarities between two given clusterings. They are most commonly used to compare results obtained from different clustering algorithms against the ground-truth clustering in the benchmark datasets.

Formally, a (disjoint) \emph{\(k\)-clustering} \(U\)of a dataset \(D\) with \(n\) data points is a partition  of \(D\) into \(k\) pairwise disjoint subsets called \emph{clusters}, \emph{i.e.}\ \(U = \{u_1, u_2, \dots, u_k\}\), with \(D = \bigsqcup_{i=1}^{k} u_i\) and for all \((i, j)\), \(u_i \cap u_j = \emptyset\) if \(i \neq j\).

CMI is a measure of the agreement between a \(k\)-clustering \(U\) and a \(r\)-clustering \(V\) of \(D\). For disjoint clusterings, it can\footnote{For a more in-depth analysis and proofs, refer to \citet{Rabbany_Zaïane_2017}.} be computed as:
\begin{equation}
    \operatorname{CMI}(U, V) = 2\frac{H(U) + H(V) - H(U, V)}{H(U) + H(V)}
\end{equation}
where \(H(U)\) is the entropy of \(U\) and \(H(U, V)\) is the joint entropy of \(U\) and \(V\).

\(H(U, V)\) and \(H(U)\) (resp. \(V\)) can be derived from the confusion matrix (a.k.a contingency table) of \(U\) and \(V\). If we consider the matrix of size \((k, r)\) whose \((i, j)\)-th element is \(n_{ij} = \abs{u_i \cap v_j}\), the size of the overlap between \(u_i\) and \(v_j\):
\begin{equation}
    \begin{array}{c|c c c c c}
             & v_1    & v_2    & \cdots & v_r\\
        \hline
         u_1 & n_{11} & n_{12} & \cdots & n_{1r}\\
         u_2 & n_{21} & n_{22} & \cdots & n_{2r}\\
         \vdots & \vdots & \vdots & \ddots & \vdots\\
         u_k & n_{k1} & n_{k2} & \cdots & n_{kr}\\
    \end{array}
\end{equation}
then
\begin{equation}
    H(U, V) = - \sum_{i=1}^{k} \sum_{j=1}^{r} \frac{n_{ij}}{n} \log\mleft(\frac{n_{ij}}{n}\mright)
\end{equation}
and, as is always true for the joint entropy, \(H(U)=H(U, U)\) and \(H(V)=H(V, V)\).

\end{document}